\documentclass[letterpaper, 10 pt, conference]{ieeeconf}  

\IEEEoverridecommandlockouts                              

\usepackage{graphics} 
\usepackage{graphicx,subfigure}
\usepackage{epsfig} 
\usepackage{amsmath} 
\usepackage{amssymb}  
\usepackage{multirow} 
\usepackage{algorithm}
\usepackage{algorithmic}
\usepackage{stfloats}
\usepackage{booktabs}
\usepackage{xcolor}
\usepackage{hyperref}

\title{\LARGE \bf
RRT* Combined with GVO for Real-time Nonholonomic Robot Navigation in Dynamic Environment
}

\author{Yuying Chen$^{1}$ and Ming Liu$^{1,2}$
\thanks{$^{1}$Yuying Chen and Ming Liu are with Department of Electronic and Computer Engineering, The Hong Kong University of Science and Technology. 
        {\tt\small ychenco@ust.hk}}%
\thanks{$^{2}$Ming Liu is with the Department of Computer Science and Engineering, The Hong Kong University of Science and Technology. 
        {\tt\small eelium@ust.hk}}%
}

\begin{document}

\maketitle
\thispagestyle{empty}
\pagestyle{empty}

\begin{abstract}

Challenges persist in nonholonomic robot navigation in dynamic environments. This paper presents a framework for such navigation based on the model of generalized velocity obstacles (GVO). The idea of velocity obstacles has been well studied and developed for obstacle avoidance since being proposed in 1998. Though it has been proved to be successful, most studies have assumed equations of motion to be linear, which limits their application to holonomic robots. In addition, more attention has been paid to the immediate reaction of robots, while advance planning has been neglected. By applying the GVO model to differential drive robots and by combining it with RRT*, we reduce the uncertainty of the robot trajectory, thus further reducing the range of concern, and save both computation time and running time. By introducing uncertainty for the dynamic obstacles with a Kalman filter, we dilute the risk of considering the obstacles as uniformly moving along a straight line and guarantee the safety. Special concern is given to path generation, including curvature check, making the generated path feasible for nonholonomic robots. We experimentally demonstrate the feasibility of the framework. 

\end{abstract}

\section{INTRODUCTION}

As robots are increasingly involved in daily life, it is common to see robots working around humans. They are assigned to a variety of tasks, from serving as tourist guides, or patrolmen to driving as autonomous cars. In most scenarios, these robots are required to navigate to target destinations with the presence of moving humans or other objects. To ensure the safety of both humans and robots and also enable robots to work efficiently, suitable control strategies applicable to the navigation tasks need to be developed. Robots are required to move towards target in a short time and avoid either static or dynamic obstacles observed by their sensors, which involves efficient path planning and valid obstacle avoidance. Though these two topics have been well researched, currently, there is no ideal solution to handling the navigation problem within cluttered dynamic environments. The typical method is treating the environment as static environment and refreshing the planning when the planned path becomes infeasible. It is simple, but inefficient as it relies on the time-consuming planning process. Navigation with the participants of pedestrians are broadly studied. However, dynamic environments that have been studied most are limited to simple environments like corridors and plazas. These environments are less cluttered and  the static obstacles are less considered and even neglected.    
 
\begin{figure}[!ht]
  \centering
    {\includegraphics[height=0.8\columnwidth]{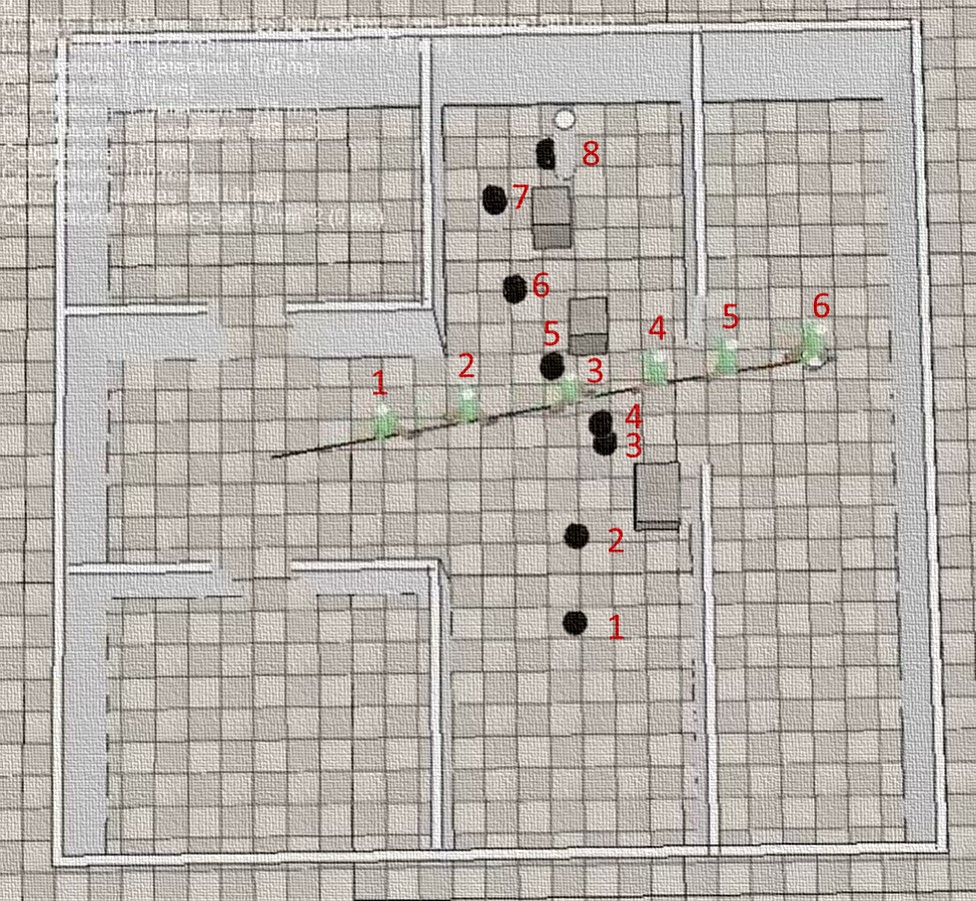} 
  \caption{Photograph of the turtlebot while navigating in a cluttered dynamic environment using the proposed scheme.
  \label{fig:figure1}} }
  \vspace{-1em}
\end{figure}

The difficulties of this kind of navigation problem are attributable to the uncertainty in the environment, particularly in the dynamic objects, and the complexity of the environment itself. As shown in Fig.\ref{fig:figure1}, the motion of the pedestrian is unknown and sensor noise exists. The environment is cluttered and no straight path is available from the start point to the goal. To handle the real-time nonholonomic robot navigation in such a dynamic cluttered environment, we propose a scheme which combines the quick path planning by RRT* and the instant collision avoidance by introducing the GVO model. The path planner reduces the collision risk along robot trajectory as the planned path bypasses the static obstacles, thus further reducing the range of concern for obstacle avoidance, saving both computation time and running time. Besides, given a reference path, the navigation problem can be solved efficiently in a more cluttered complex environment. Furthermore, by introducing uncertainty for the dynamic obstacles with a Kalman filter, we reduce the risk of considering obstacles as moving along a straight line with a consistent speed.  

The rest of the paper is organized as follows. Related work is covered in Section II. In Section III, we firstly present the framework of the navigation system. Then each part of the system is introduced in detail, including the path generator, path follower and obstacle avoidance. In Section IV, the experimental platforms are introduced and the results analyzed and evaluated. Finally, conclusions are drawn in Section V. 

\section{Related works}
Existing works that target solving the complex obstacle avoidance problem can be roughly classified into two categories, namely model-based and learning based approaches. As the study of deep learning continues to heat up, learning based approaches \cite{tai2017virtual},\cite{chen2017socially} have been put forward. The key idea is to mimic or learn human decision policies toward solving the complex navigation problem. The training process is time- and data-consuming and the performance of these methods may degrade a lot if the robot is put into a new environment. Different from the learning based methods, model-based methods rely on reasonable geometric rules or potential field are considered more computationally efficient. In 1998, one of the most representative works,\cite{fiorini1998motion}, proposed the velocity obstacles (VO) model, which utilizes collision cones to define the region of velocity that will cause a collision at some time in the future. Keeping the velocity selection outside the cone ensures safe navigation.  Since 1998, variations of VO have been proposed to solve problems encountered in different scenarios. As the VO model holds the assumption that dynamic obstacles move passively and will not react to a robot, which is not true for multiple agents, the reciprocal velocity obstacles(RVO)\cite{van2008reciprocal} method modifies the position of the collision cone by assuming every robot shares half of the responsibility of collision avoidance. And optimal reciprocal collision avoidance(ORCA)\cite{van2011reciprocal} defines the set of safe velocities to be a half plane with respect to VO and guaranteed local-free motion of a large number of robots. As the above methods are limited to holonomic robots, various approaches have been proposed to extend them to differential-drive\cite{alonso2013optimal},\cite{snape2010smooth} and car-like robots\cite{alonso2012reciprocal}. To be more general, Wilkie \textit{et.al.}\cite{wilkie2009generalized} defined the velocity cone as generalized velocity obstacles (GVO) and made it general and applicable for robots with different kinematic constraints. For the methods mentioned above, the precise status, including the location and velocity of the moving obstacles is required and only the instant velocity is considered, which make these methods lack of anti-interference capability for real environment applications. Another proverbial approach for obstacle avoidance is the social force model \cite{helbing1995social}, which mainly focuses on the interactions among various agents and defines the attractive force and repulsive force for navigating toward the goal and avoiding the obstacles, respectively. However, it requires the knowledge of final destination of every agent.

Generally, path planning aims at finding a curve starting from a start node, to the target. And we are dealing with a local path planning problem with a near target and are required to give a path with high resolution. Various methods have been proposed, and can be divided into complete and probabilistically complete algorithms. Of probabilistically complete approaches, i.e., sampling based methods, the most representative methods are rapidly-exploring random trees(RRT)\cite{lavalle1998rapidly} and probabilistic road maps(PRM)\cite{kavraki1996probabilistic}. One of the most famous variations of the RRT is RRT*\cite{karaman2011sampling}, which adds an extra "rewiring" step to the RRT tree and converges towards an optimal solution. Other methods that guarantee optimality are based on graph search and are called complete algorithms.　A*\cite{hart1968formal}, which combines best-first search and Dijkstra's algorithm\cite{dijkstra1959note} to find the optimal solution by searching among all possible paths. Dynamic A* search(D*)\cite{stentz1994optimal} focuses on the updates of cost to minimize state expansions and further reduces computational costs. Other methods, including some local planning algorithms, are represented by the dynamic window approach(DWA)\cite{fox1997dynamic} and vertical field histogram(VFH)\cite{borenstein1991vector}. The DWA generates acceptable velocity by sampling and evaluates traces by heading angle error. As it treats all obstacles equally and shows no concern for the motion state of obstacles, it is limited to applications in a static environment.
\section{System}
\subsection{Framework}
\begin{figure}[!ht]
  \centering
    {\includegraphics[width=1\columnwidth]{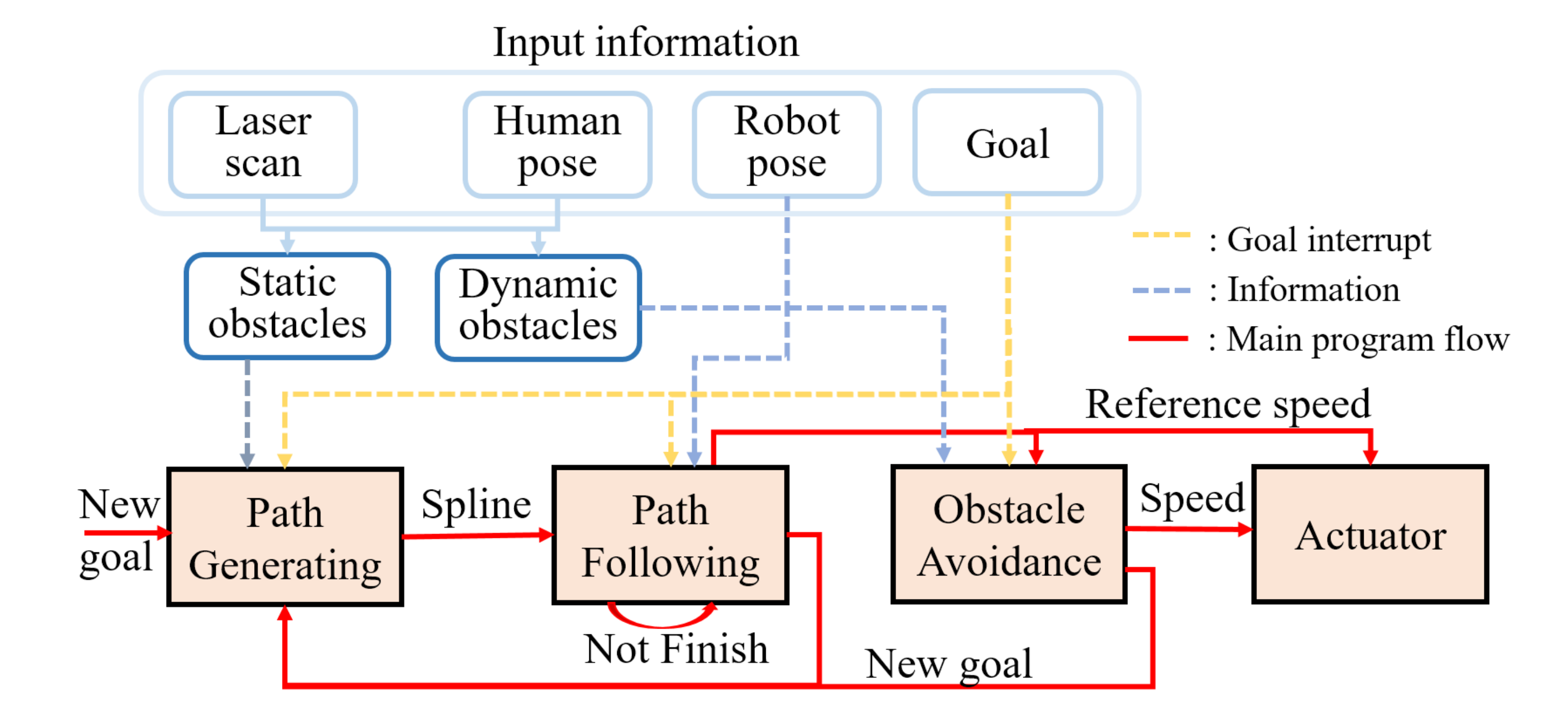} 
  \caption{Framework of the system.
  \label{fig:framework}} }
  \vspace{-1em}
\end{figure}

The whole system consists of robot trajectory generation, dynamic object status estimation, obstacle avoidance and path-following control. At the beginning, objects defined by laser points are clustered and classified into static and dynamic obstacles. Only static points are considered for path generation. Then a collision free and smooth path is generated by RRT* and spline interpolation. After that, laser point clustering of humans is registered and the positions are treated as the observation input of a Kalman filter. The Kalman filter is updated every cycle and both moving speed and position can be directly obtained. If the obstacles are far away from the robot, a path-following controller will take control and the robot will follow the generated path. Otherwise, obstacle avoidance will dominate and the controller will only offer a reference action. All the obstacles are treated as GVOs. By estimating the status of the robot and obstacles after taking a certain action within a certain time, a set of actions that satisfy the collision condition can be acquired. The final action is decided by the similarity to the reference velocity. The framework is shown in Fig.\ref{fig:framework}.

\vspace{-0em}

\subsection{Path generator}
\begin{figure}[!ht]
  \centering
    {\includegraphics[height=0.35\columnwidth]{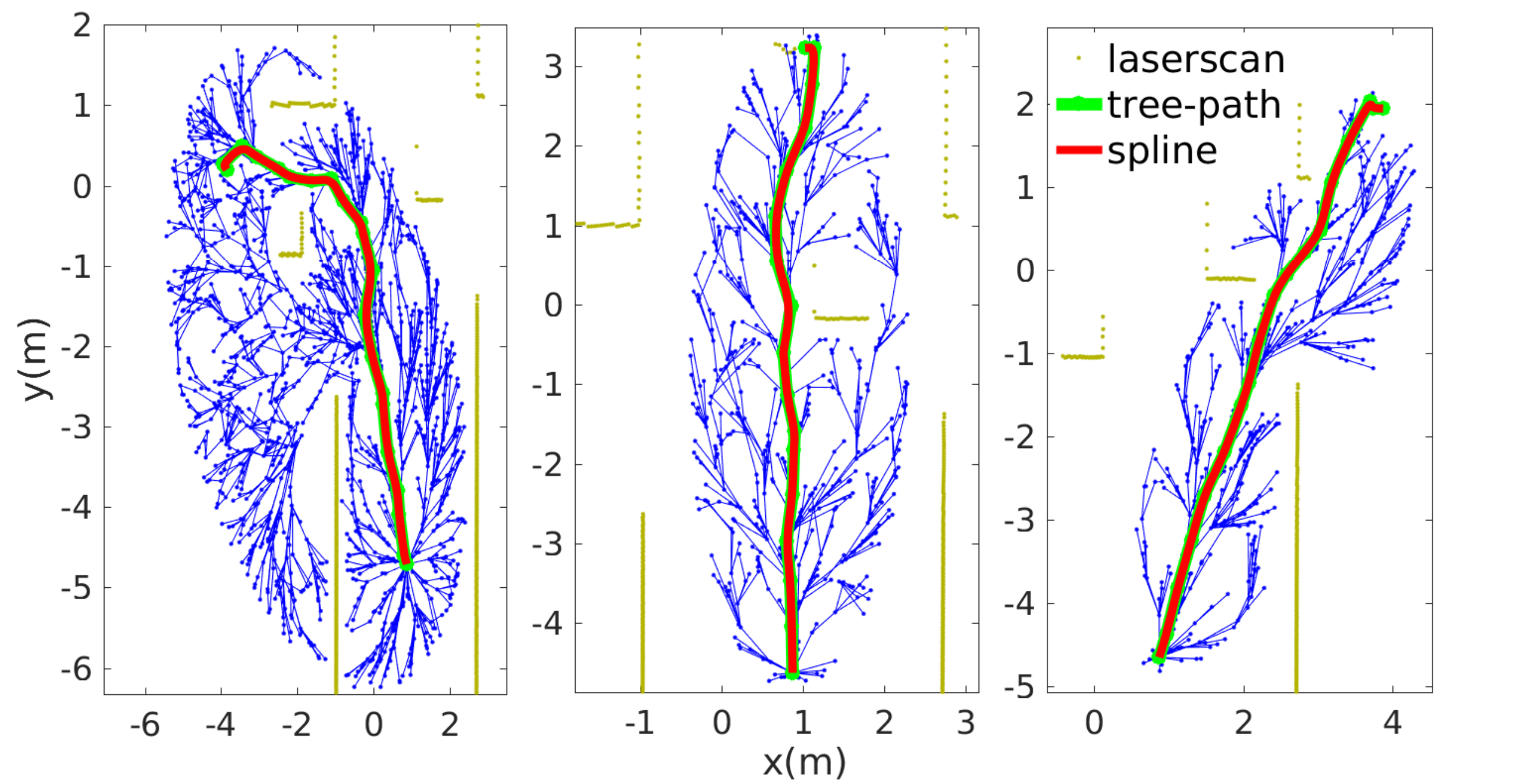} 
  \caption{Trace examples generated by informed-RRT* and spline interpolation. To generate the tree, the sampling density is set to be 8 $nodes/m^2$ and the largest distance between two nodes is 0.6m. And there are no iterations for time consideration.
  \label{fig:path}} }
  \vspace{-1em}
\end{figure}

The path generator is mainly based on the RRT* algorithm with extra constraints added to obtain satisfactory solutions. As shown in Algorithm 1, to begin with, static obstacles are extracted by DBSCAN clustering because only static obstacles are considered for path generation. Different from RRT*, informed RRT* \cite{gammell2014informed} induces heuristic-biased sampling, which increases the sampling probability inside the heuristic sampling domain while reducing the probability outside. For informed RRT*, the heuristic domain is an ellipse, with its shape defined by the distance between the start point and goal as well as the given minimal distance cost. Here, the ellipse is defined in the same way, but all the samplings are done inside the ellipse as we get the best cost from the standard RRT* algorithm and ascertain that one solution can be obtained inside it. The sampling density of the RRT* is set to be much smaller than the latter. Besides the sampling method, a collision check and curvature check are introduced to better satisfy the motion conditions of nonholonomic robots. After finding the path to goal, spline interpolation is conducted to get a smooth path. Examples of generated paths are shown in Fig.\ref{fig:path}

\begin{algorithm}[!ht]
	
    \caption{Path generator}
    \label{alg:addpg}
    \begin{algorithmic}[0]   
    \STATE{
    Extract static $Ob_{s}$ and dynamic obstacles $Ob_{d}$. \\
    Generate path by RRT* with low sampling density and take the distance cost as the minimal cost obtained $c_{best}$.\\
    Assume the robot position as $P_{robot}=(P_{r_x},P_{r_y})$, goal position as $P_{goal}=(P_{g_x},P_{g_y})$,and the ideal minimal distance as
    $c_{min}=\sqrt{(P_{g_x}-P_{r_x})^2+(P_{g_y}-P_{r_y})^2}$. 
    }
    \FOR {$num_{nodes}  <= area*density_{nodes}$}
    \STATE{
    Generate potential node $A$ by sampling inside an ellipse:
    \begin{align*}
  \begin{bmatrix}
  x\\
  y
  \end{bmatrix}
  &= 
\begin{bmatrix}
\frac{ P_{g_x}-P_{r_x} }{ c_{min} } & - \frac{ P_{g_y}-P_{r_y} }{ c_{min} } \\
 \frac{ P_{g_y}-P_{r_y} }{ c_{min} } & \frac{ P_{g_x}-P_{r_x} }{ c_{min} } 
\end{bmatrix}
\begin{bmatrix}
\frac{ c_{best} }{ 2 } & 0 \\
 0 & \frac{ \sqrt{c_{best}^2-c_{min}^2} }{ 2 } 
\end{bmatrix}
\\
&\cdot
\begin{bmatrix}
x_0 \\
 y_0
\end{bmatrix}
+
\begin{bmatrix}
\frac{ P_{g_x}+ P_{r_x}}{ 2 } \\
 \frac{ P_{g_y}+ P_{r_y}}{ 2 }
\end{bmatrix}
,where \quad x_0^2 +y_0^2 =1
    \end{align*}
    Adjust $A$ to ensure that $A$ is close to at least one node in the accepted node set.\\
    Get the parent node of $A$, marked as node $B$ and the parent node of $B$, marked as node $C$.\\
    \IF{ $min(dis(AB,Ob_{s}) > dis_{th} $ and $|angle(AB)-angle(BC)|<angle_{th}$ } \STATE {Add new node $A$.} \ENDIF
    }
    \STATE{Do curvature check and collision check during rewiring.}
    \ENDFOR\\
    \STATE{Track back from the closest node to goal to start point}
    \STATE{Generate trace points by spline interpolation}
\end{algorithmic}
\end{algorithm}

\vspace{-0em}

\subsection{Path follower}
\begin{figure}[!ht]
  \centering
    {\includegraphics[height=0.45\columnwidth]{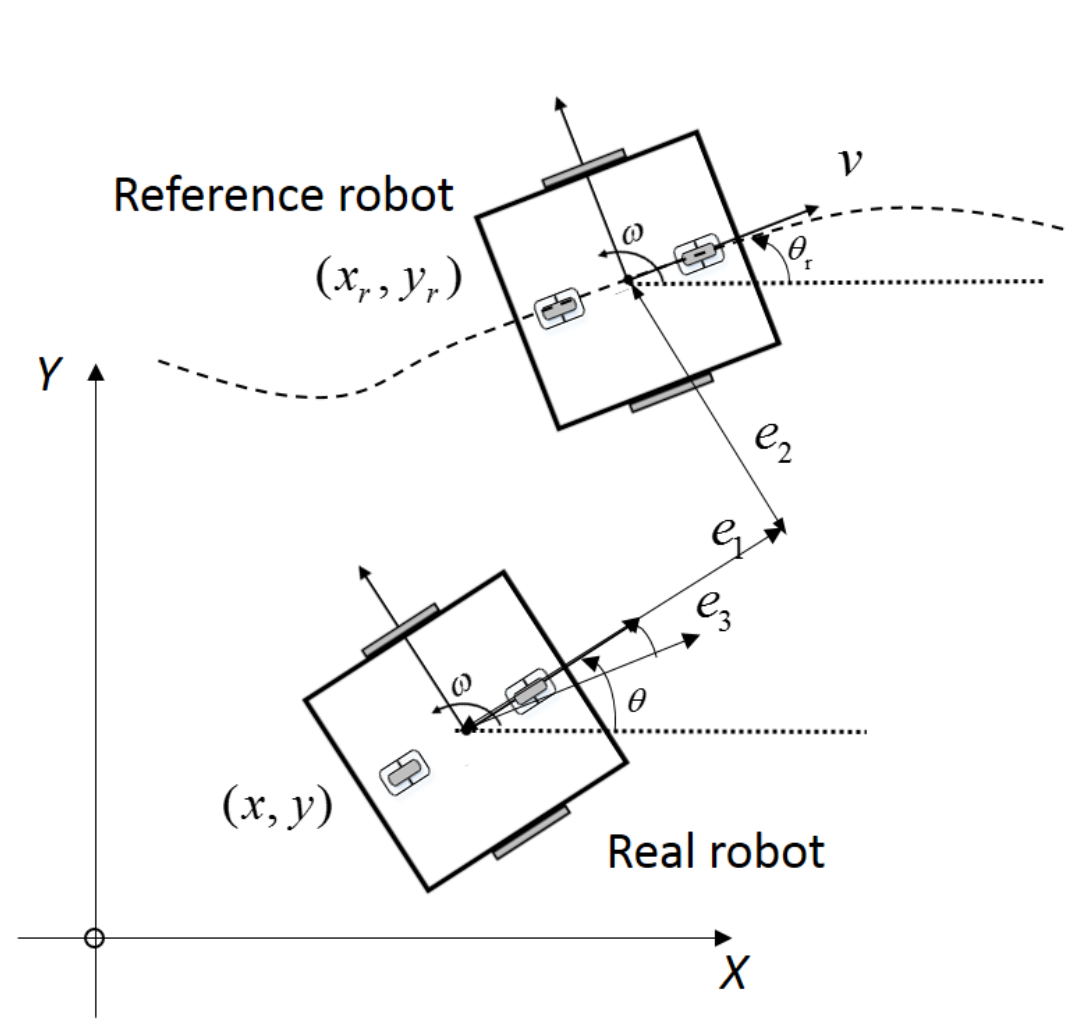} 
  \caption{Differential drive robot model and error definition for closed loop control.$e_1,e_2$ and $e_3$ are defined in the real robot frame, and poses of robots are defined in the world frame.
  \label{fig:model}} }
  \vspace{-1em}
\end{figure}


\begin{align}
\begin{bmatrix}
 \dot{x} \\ 
 \dot{y}\\ 
 \dot{\theta}
\end{bmatrix}
&=
\begin{bmatrix}
cos\theta & 0 \\
sin\theta & 0 \\
 0 & 1
\end{bmatrix}
 \cdot
\begin{bmatrix}
 v \\ 
 \omega
\end{bmatrix}  \label{eq:1}\\
\begin{bmatrix}
 e_1 \\ 
 e_2 \\ 
 e_3
\end{bmatrix}
&=
\begin{bmatrix}
-cos\theta & -sin\theta & 0 \\
sin\theta & -cos\theta &0 \\
 0 & 0 & -1
\end{bmatrix}
 \cdot
\begin{bmatrix}
 x-x_r \\ 
 y-y_r \\ 
 \theta - \theta_r
\end{bmatrix} \label{eq:2}\\
\begin{bmatrix}
 u_{e1} \\ 
 u_{e2} 
\end{bmatrix}
&=
\begin{bmatrix}
-k_1 & 0 & 0 \\
0 & -sign(u_{r1})k_2 &-k_3
\end{bmatrix}
 \cdot
\begin{bmatrix}
 e_1 \\ 
 e_2 \\ 
 e_3
\end{bmatrix} \label{eq:3}\\
\begin{bmatrix}
 v \\ 
 \omega 
\end{bmatrix}
&=
\begin{bmatrix}
cose_3 & 0 \\
0 & 1
\end{bmatrix}
 \cdot
\begin{bmatrix}
 u_{r1} \\ 
 u_{r2}
\end{bmatrix}
-
\begin{bmatrix}
 u_{e1} \\ 
 u_{e2}
\end{bmatrix} \label{eq:4}\\
k_1&=k_3=2\xi\sqrt{u_{r2}(t)^2 +g u_{r1}(t)^2} \label{eq:5}\\
k_2&=g\cdot|u_{r1}| \label{eq:6}
\end{align}

After path generation, trace is passed to a path follower. We use the closed loop controller proposed by Klancar \textit{et al.}\cite{klancar2005mobile} to make the robot move along a reference path. The robot architecture can be seen in Fig.\ref{fig:model}. For a differential drive robot, the motion equations are described by Eq.\ref{eq:1}, where $v$ and $\omega$ are the forward and angular velocities, and $\theta$ is the forward direction of the robot in the world frame. The error between the real pose $(x,y,z)$ and the reference pose $(x_r,y_r,\theta_r)$ in the frame of the real robot can be calculated by Eq.\ref{eq:2}. Multiplying the error by gain matrix $K$, we can get the feedback $(u_{e_1},u_{e_2})$ (shown in Eq.\ref{eq:3}). The final output actions$(u_1,u_2)$ can be obtained from reference actions $(u_{r_1},u_{r_2})$ and $(u_{e_1},u_{e_2})$. The $K$ matrix depends on the reference actions. And $\xi$ and $g$ have a large influence on the result. In experiments, we found that with large g values, the robot will move in a zigzag, as the controller becomes too sensitive to the error. In this case, the error is defined as the closest distance from the current robot position to the spline, and the reference actions are given by looking several steps forward. If all the obstacles are out of collision range, this controller will take control and send the velocity command $(v,\omega)$. Otherwise, the output of the controller will serve as reference actions and be passed to the obstacle avoidance.

\subsection{Obstacle avoidance}
For obstacles avoidance, we learn from the GVO model and apply it to the differential drive robot. The GVO model was proposed to solve the real-time navigation problem in dynamic environments with car-like robots. The key idea is to find the acceptable actions that will avoid collision in the near future. Different from most of the VO models, the GVO model has no requirement of linear motion of the robots, which makes it convenient to extend to nonholonomic robots. Although the paper only focuses on dynamic obstacles, in theory, it is also applicable to static environment, which make it usable for applications in complex environments. As shown in Algorithm 3, firstly, the laser data are divided into dynamic and static obstacles. Provided with a reference path, the range of static obstacles considered can be suppressed. After getting the human pose as an observation, the Kalman filter is updated and then both the position and velocity of humans $[p_x,p_y,v_x,v_y]^T$ can be obtained. Given a sampling space, which is mainly confined by the maximum forward velocity and angular velocity of the robot, one potential action is generated. Then robot poses at time $t$ can be derived, as shown in Eq.\ref{eq:7} and Eq.\ref{eq:8}. Different obstacles are handled differently. The relative position of static obstacles $P_{ob_s}(t)$ is certain if the error of the robot's position can be omitted and time threshold $t_{s_th}$ is small (will not induce a large odom error). So the minimal distance between the robot and obstacles given $t\in[0,t_{s_{th}}]$ can be derived easily. As there exist both pose uncertainty and velocity uncertainty, the motion of humans cannot be reduced to a simple linear motion model. In this paper, human pose at time t, $P_{human}(t)$, is treated as a sum of two Gaussians and is also normally distributed with a distribution of $N(\mu_p+\mu_vt,\Sigma_p + t^2\Sigma_v)$. This is straightforward as a long time will increase the uncertainty of the predicted human pose. To ensure the safety of humans, we set a threshold and when the probability that the robot will collide with a human goes high enough, the time is recorded and the action is rejected. After sampling n times, we get two sets. If the accepted actions set is not empty, the difference between desired actions and proposed actions will be the rule of choosing the final action. The most common one is the 2-norm, which was also used in \cite{wilkie2009generalized}.  If there is no good choice, instead of stopping and waiting for the next loop, the corresponding actions with maximum time will be chosen, and when the time to collide is less than $t_{c_{th}}$, the robot will stop. All the thresholds in this model depend on the kinematic constraints of the robot:
\begin{eqnarray}
&x(t)= \frac{v}{\omega}sin(\theta +\omega t) - \frac{v}{\omega}sin(\theta)  \label{eq:7}\\
&y(t)= -\frac{v}{\omega}cos(\theta +\omega t) + \frac{v}{\omega}cos(\theta).  \label{eq:8}
\end{eqnarray}
\begin{algorithm}[!ht]
    \caption{GVO model}
    \label{alg:addpg}
    \begin{algorithmic}[0]   
    \STATE{
    Get desired actions $u^{*}$ from path follower\\
    Classify laser points into $Ob_{s}$  and $Ob_{d}$}\\
    \STATE{Update Kalman filter and get estimated human pose $(p_x,p_y)$ and velocity $(v_x,v_y)$}\\
    \STATE{Get static obstacles $\{Ob_{s_1},Ob_{s_2}...\}$ within distance $dis_{sta}$ }\\
     \FOR{i = 0 to n}
     \STATE{$free=True$}
     \STATE{Sample one action $u=(w,v)$ from action space}
     \STATE{Get estimated robot pos $P_{robot}(t)$ at t}
     \FOR{all dynamic obstacles $Ob_d$ and static obstacles $Ob_s$}
     \STATE{Let $D_s(t)$ be the distance between $Ob_s$ and robot at time $t$, $t<t_{s_{th}}$}
	 \STATE{$t_{min}=min(argmin(D_s(t)),t_{min})$}
	 \IF{ $D_s(t_{min}) < radius_{robot}$ } \STATE {$free=False$} \ENDIF
	 \STATE{Let $f_d(t)$ be the normalized PDF value of human position distribution at $P_{robot}(t)$, $t<t_{d_{th}}$}
	 \WHILE{$t<t_{d_{th}}$} 
	 \IF{ $f_d(t) > p_{th}$ } \STATE {$free=False$\\$t_{min}=min(t_{min},t)$\\break} \ENDIF
	 \STATE{$t=t+\delta t$}
	 \ENDWHILE	 
	 \ENDFOR\\  
	 \IF{ free } \STATE {Add $(w,v)$ to accept actions set $\mathbb{A}$}     \ELSE \STATE{Add $(w,v)$ to reject actions set $\mathbb{R}$}\ENDIF
     \ENDFOR \\
     \IF{ $\mathbb{A} \neq \emptyset$  } \STATE {$u = argmin(f(u-u^{*}))$}     \ELSE \STATE{$u = argmax(t_{min}(u)),u\in \mathbb{R}$ \\
     \IF{ $t_{min} < t_{c_{th}}$ } \STATE {$u = 0$}     
      \ENDIF  
     }\ENDIF
     \STATE{Return $u$}

\end{algorithmic}
\end{algorithm}

\section{Experimental results}

\subsection{Platform}
\begin{figure}[!ht]
  \centering
     \subfigure[platform]{\includegraphics[height=0.32\columnwidth]{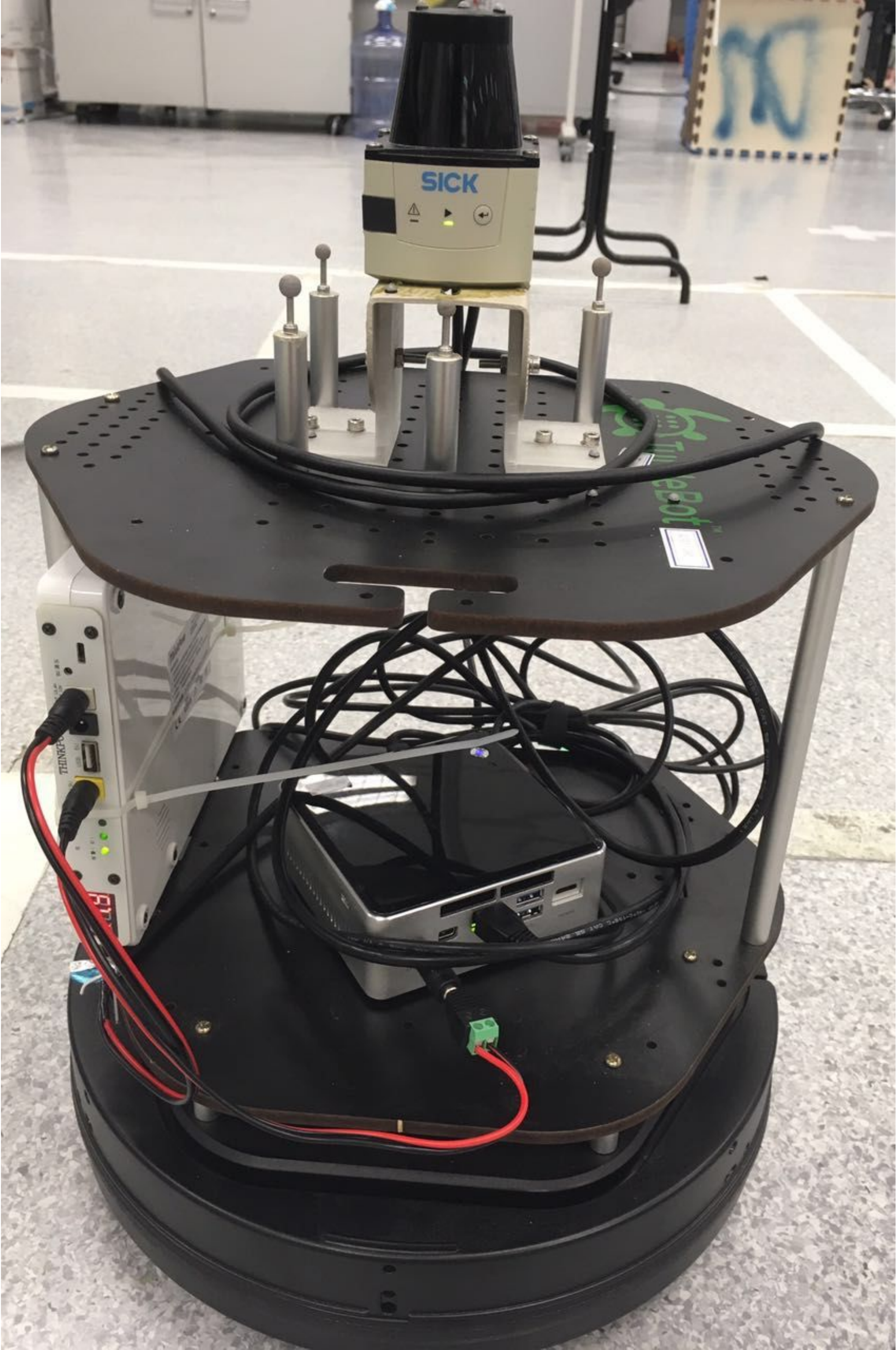}
    \label{fig:platform}}
      \subfigure[test scene]{\includegraphics[height=0.32\columnwidth]{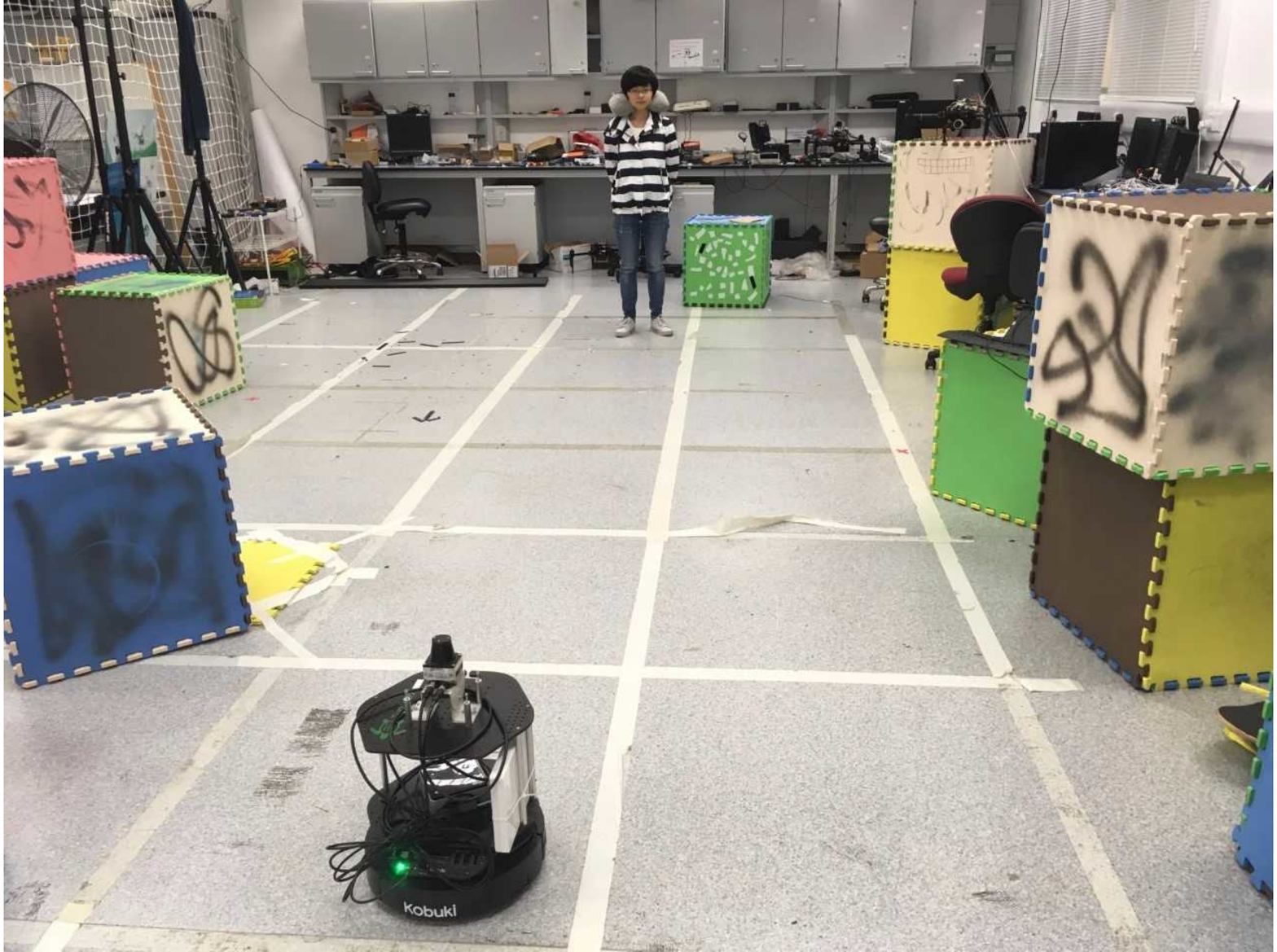}
    \label{fig:real}} 
  \caption{(a) Test platform and (b) test environment. To use the motion capture system, ball reflectors were attached to the platform and walking man for localization. (b) shows the test scene for dynamic obstacle avoidance.}
    \vspace{-0.5em}
\end{figure}

To show the performance of the planner, navigation in both a virtual and a real environment was tested. The virtual environment was built in V-rep\cite{rohmer2013v}. Walls and blocks were placed as static obstacles and a walking man was introduced as a dynamic obstacle. The man walked towards a random target with simple path planning and could not avoid obstacles in a timely way. In both environments, we used \textit{Turtlebot} as the mobile ground platform, equipped with a SICK TiM561\footnote{https://www.sick.com/de/en/detection-and-ranging-solutions/2d-lidar-sensors/tim5xx/tim561-2050101/p/p369446} 2D laser range finder(LRF). The experimental configuration is shown in Table \ref{tab:config}. To capture the pose of the human, we tested the whole system inside a motion capture system, OptiTrack\footnote{http://optitrack.com/products/prime-41/}.The test platform and scene are shown in Fig.\ref{fig:platform} and Fig.\ref{fig:real}. 
\begin{table}[h]
\caption{Test configuration}

\label{tab:config}
\begin{center}
  \vspace{-2em}
\begin{tabular}{|c|c|c|}
\hline
Components & Parameters & Comments\\
\hline
Turtlebot & $v_{max} = 1.6m/s, \omega_{max}=\pi rad/s$ & Kobuki\\
\hline
SICK 2D LRF & FOV=$270^{\circ}$, angle resolution= $0.33^{\circ}$, &  can be used\\  & range=[0.05m,10m], $scan_{feq}$=15HZ & outdoors\\
\hline
Computer&Intel NUC Kit NUC5i7RYH & - \\  
\hline
Lithium battery & 12 V and 19V output & Power supply \\
\hline
\end{tabular}
\end{center}
\end{table}

\subsection{Evaluation}
\begin{figure}[!ht]
  \centering
     \subfigure[test scene1]{\includegraphics[height=0.32\columnwidth]{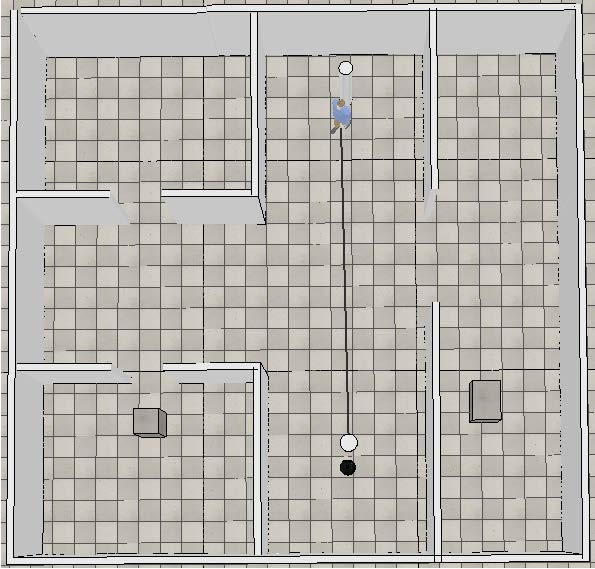}
    }
      \subfigure[test scene2]{\includegraphics[height=0.32\columnwidth]{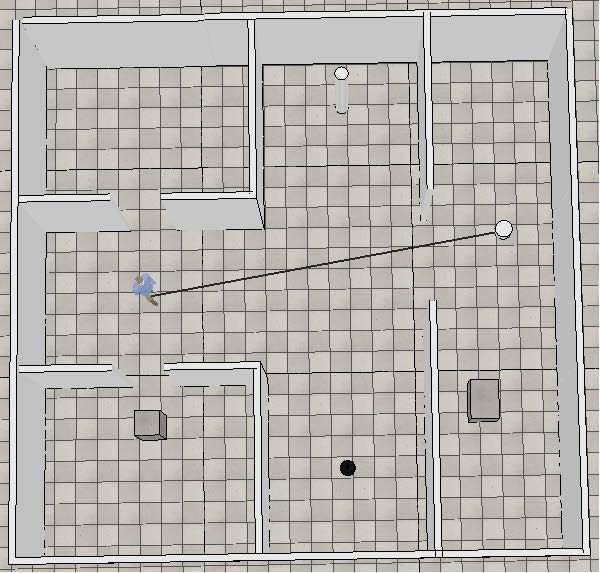}
    } 
     \subfigure[test scene3]
{\includegraphics[height=0.32\columnwidth]{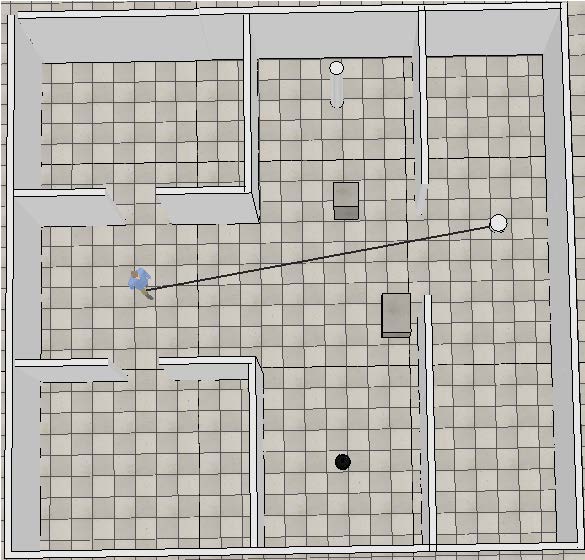}
    }
      \subfigure[test scene4]{\includegraphics[height=0.32\columnwidth]{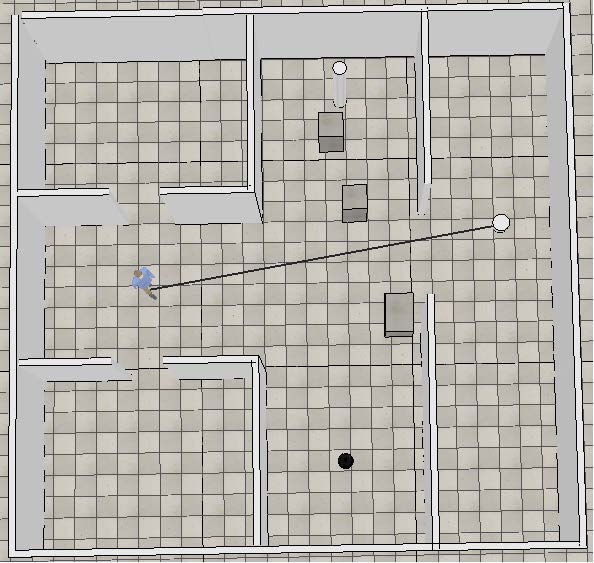}
   }     
    
  \caption{Test secenes in virtual environment.\label{fig:test}}
    \vspace{-0.5em}
\end{figure}

To measure the performance of the navigation system, we compared the performance of the proposed method with the GVO model without path planning in terms of average navigation time, success rate, etc. We aimed to evaluate the effect of including a path generator, as in most cases, the VO model has no plan for velocity control with its reference velocity simply set towards the target. Here, the compared GVO model was combined with state estimation of moving obstacles and the same controller for velocity control. Four different scenes were evaluated for the virtual environment. For better visualization, both the goal of the robot and human were shown as columns. And to test the robustness of the scheme, extreme cases where the human would definitely collide with the robot if no strategy was adopted were tested, including a scene where both the start point and goal of the human and robot were in the same line, and the cross scenario. The complexity of the task was increased from scene1 to scene4(shown in Fig.\ref{fig:test}). We ran the testing 10 times for each scene with the two different methods. We also tested the algorithm in a real environment and recorded the robot trace for analysis.

\subsection{Results}

\begin{figure}[!htb]
  \centering
     \subfigure[2D-trace]{\includegraphics[height=0.5\columnwidth]{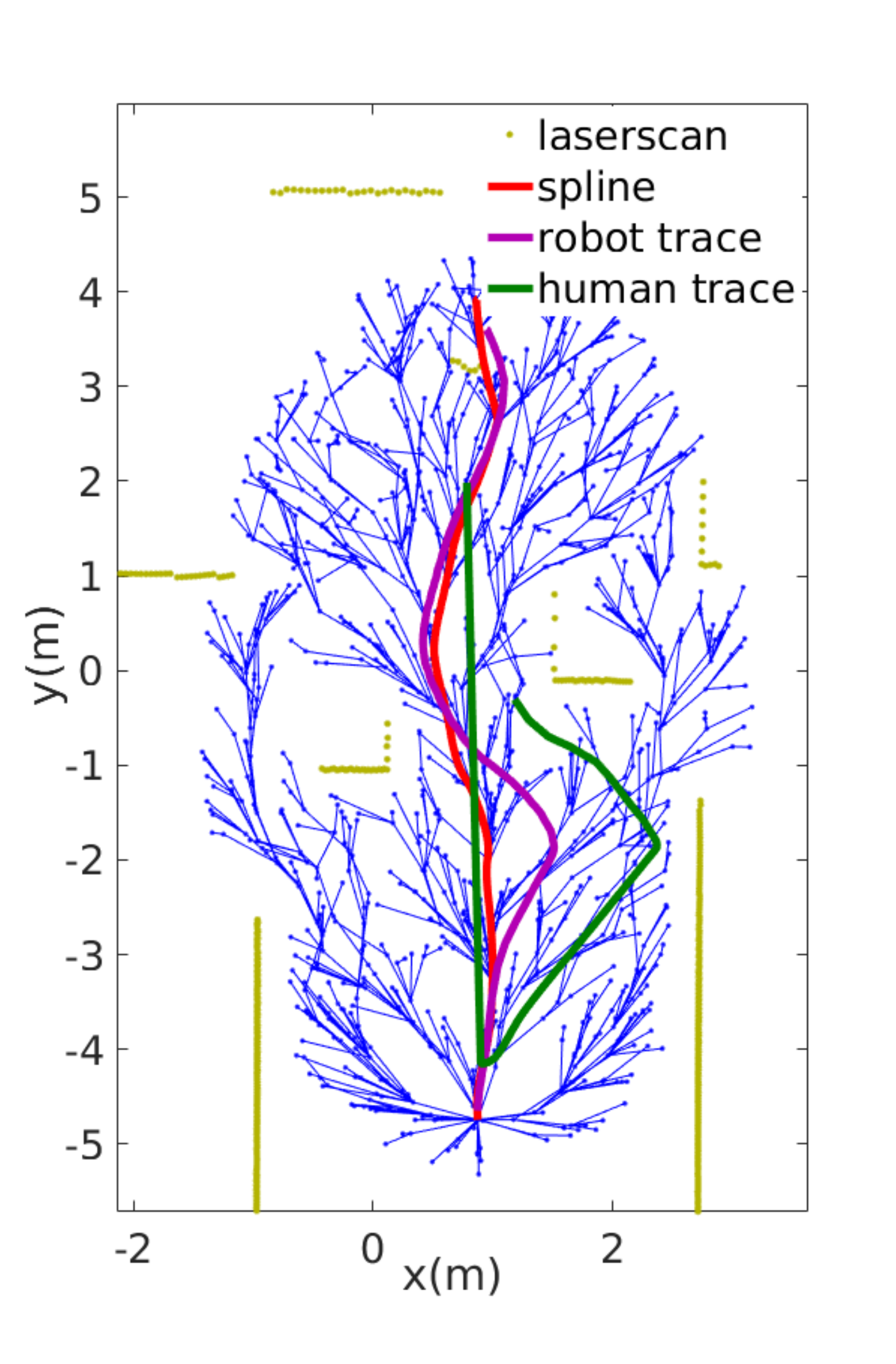}
    \label{fig:trace}}
      \subfigure[3D-trace]{\includegraphics[height=0.42\columnwidth]{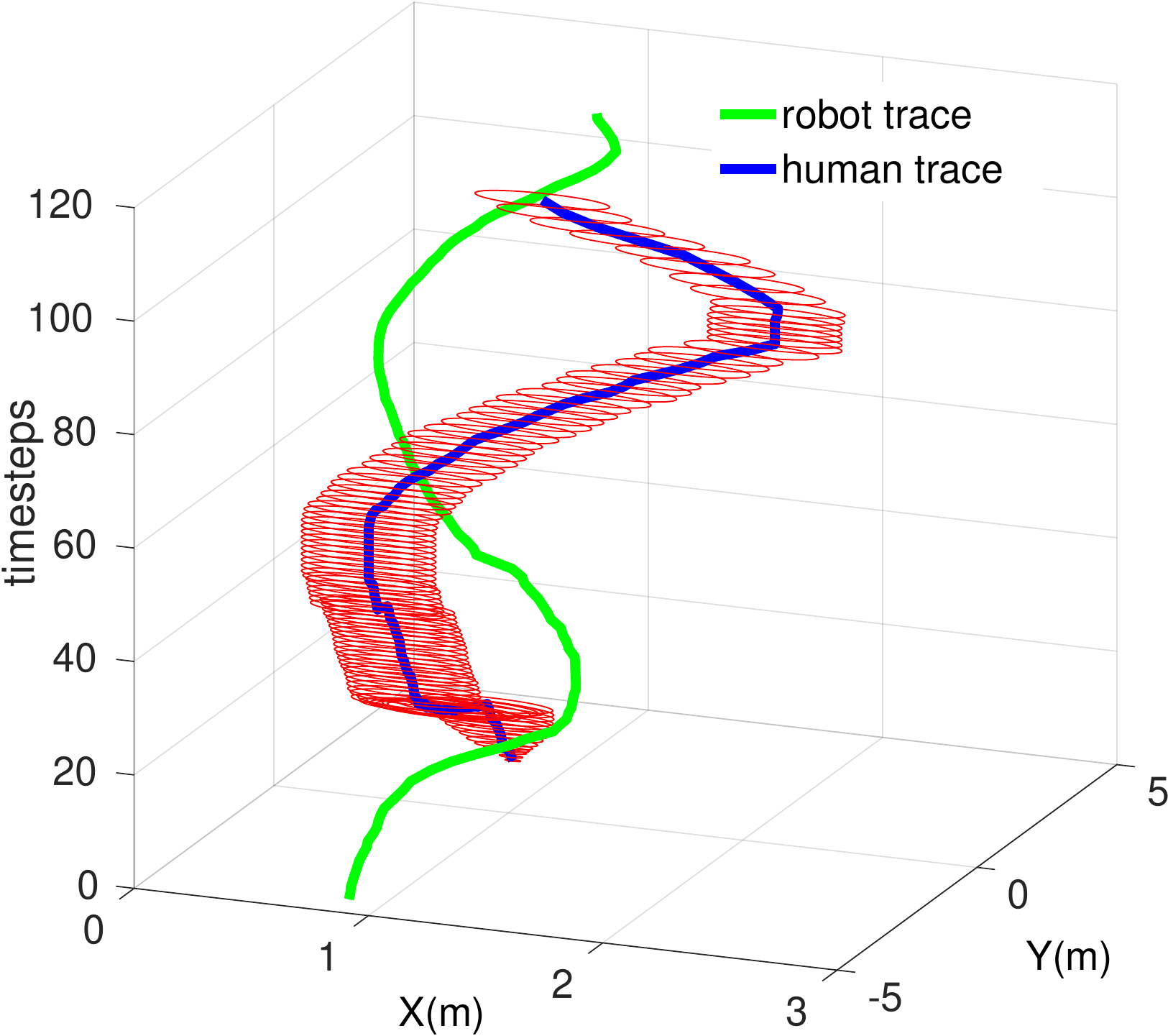}
    \label{fig:trace1}} 
  \caption{Generated and real trace of a robot during navigation. In Fig.\ref{fig:trace}, the red line is the spline, and the magenta and green lines are the trajectories of the robot and a human respectively.  Fig.\ref{fig:trace1} shows the robot trace and human trace with time steps; the red circles are error ellipses at one $\sigma$ }
  \vspace{-1em}
\end{figure}

\begin{figure}[!ht]
  \centering
     \subfigure[Success rate]{\includegraphics[height=0.35\columnwidth]{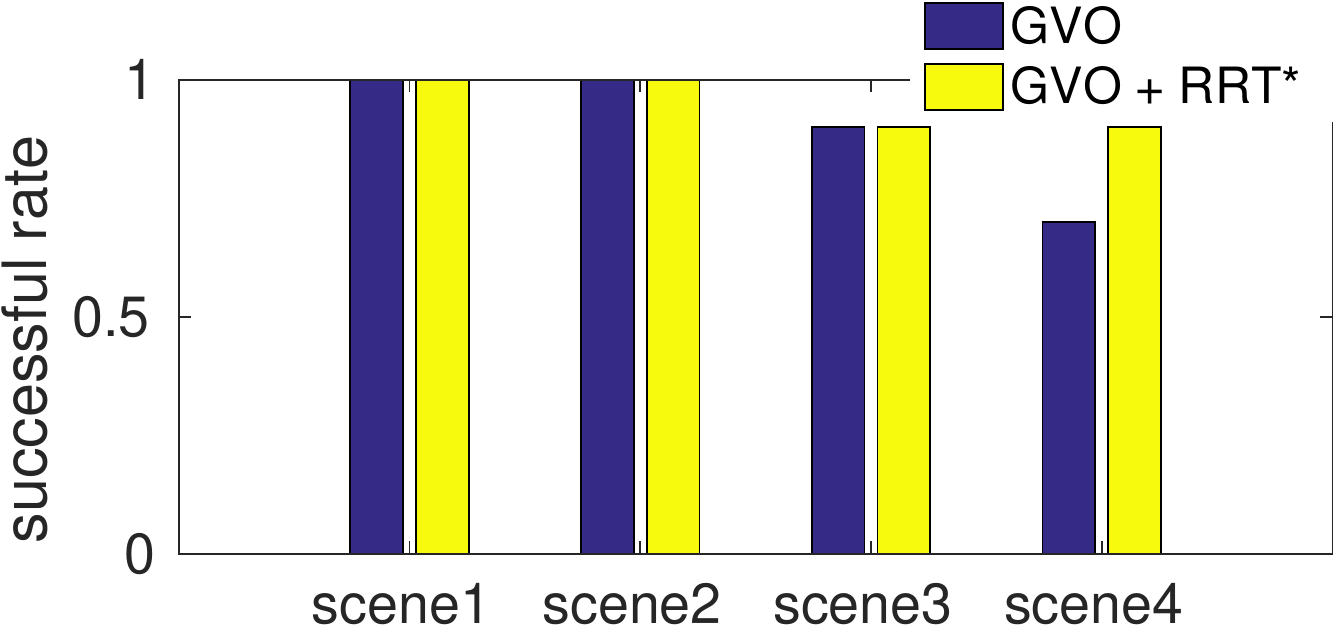}        \label{fig:result1}} 
      \subfigure[Average time]{\includegraphics[height=0.32\columnwidth]{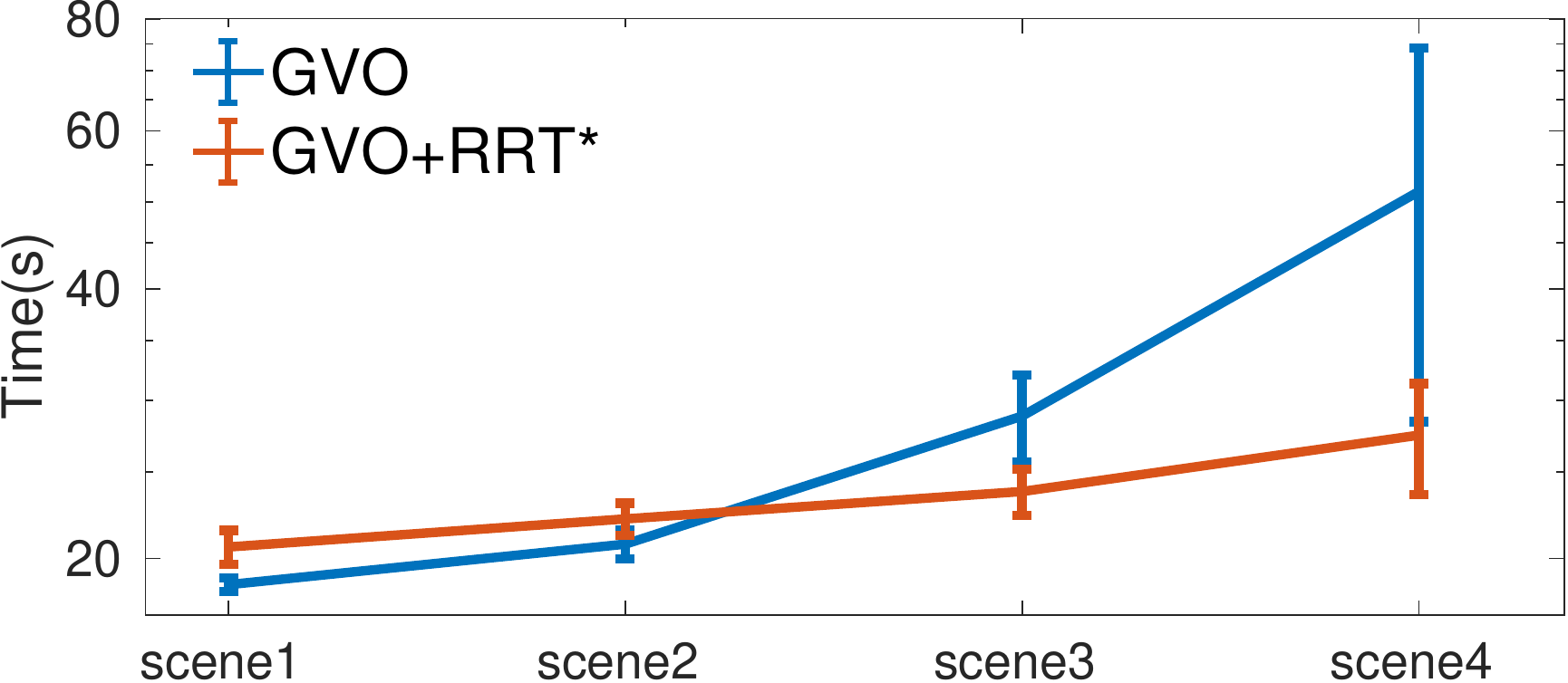} 
\label{fig:result2}}    
  \caption{(a)Success rate(b)Average time spent in different scenes for GVO only and GVO+RRT*.
 }
  \vspace{-1em}
\end{figure}

\begin{figure*}[!ht]
  \centering
    {\includegraphics[height=0.5\columnwidth]{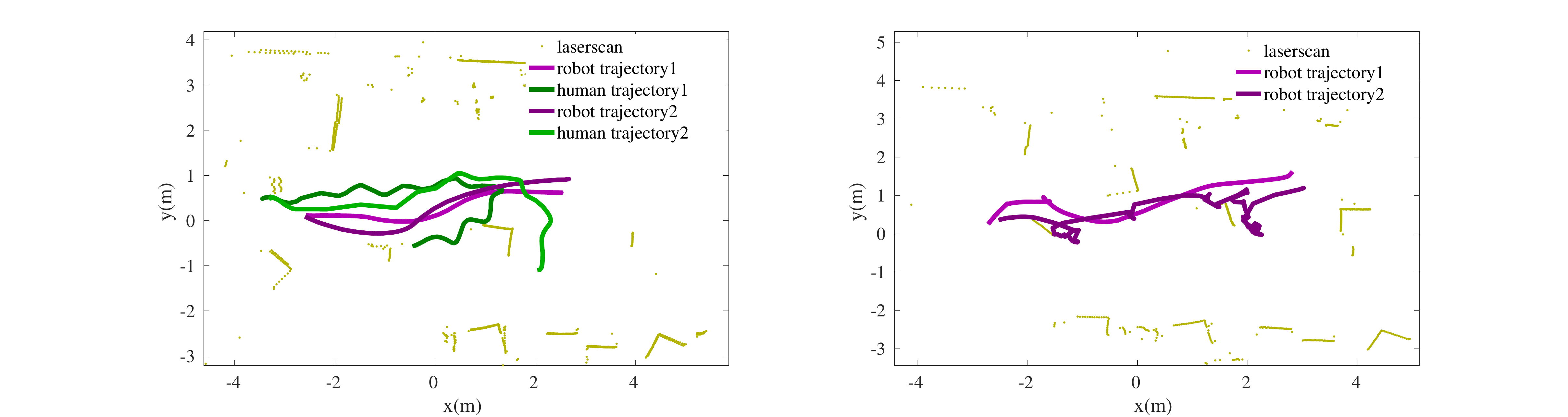} 
  \caption{Robot navigation test in real environment. Robot trajectory1 and Robot trajectory2 stand for robot navigation based on GVO + RRT* and GVO alone, respectively. Human trajectory1 and human trajectory2 represent the human activity during the navigation, respectively.
  \label{fig:realtest}} }  
\end{figure*}

A successful demonstration of the robot navigation through the complex environment is shown in Fig.\ref{fig:trace} and Fig.\ref{fig:trace1}. Due to the limit of optimization time, the planned path(red) of the robot is suboptimal. The robot follows the path before the awareness of the possible collision caused by the human coming right ahead. And it goes to the right side to avoid the human and return to the original path after the potential collision being resolved. The human trace is represented by a series of Gaussian distributions provided by the Kalman filter.  The test results in the simulation environment are shown in Fig.\ref{fig:result1} and Fig.\ref{fig:result2}. The bar graph in Fig.\ref{fig:result1} shows that both GVO and GVO combined with RRT* finished the task successfully in a dynamic environment given state estimation of the human by a Kalman filter. Increasing the complexity of the static scene will greatly influence the performance of the GVO model. As is shown in Fig.\ref{fig:result2}, both finishing time and time fluctuation increased. Though RRT* took some time to generate a path, which made the time a little bit longer in simple environments, it was stable when increasing the number of obstacles. Since the human in the simulation environment could not react to a potential collision immediately, the results are considered to have shown the worst situation. 
 We also test the strategy in a real environment. As shown in Fig.\ref{fig:realtest}, the trajectories were smooth in a simple dynamic environment. When more static obstacles appeared, GVO+RRT* kept a smooth trajectory, while the robot controlled by GVO rotated and changed direction to avoid collision. 

\section{CONCLUSIONS}
In this paper, we have demonstrated a scheme for real-time nonholonomic robot navigation in dynamic environments, which combines RRT* and GVO to deal with path planning and obstacle avoidance. We also introduce a Kalman filter to model human motion. The proposed navigation scheme was proved to be more robust to complicated environments than GVO alone. 

Future work includes integrating human 3D pose estimation from RGB images or 3D lidar. As it is applicable to car-like robots, further extending our scheme to autonomous driving is promising and  would have great significance.





\
\
\bibliographystyle{IEEEtran}
\bibliography{navibib}

\begin{thebibliography}{10}
\providecommand{\url}[1]{#1}
\csname url@samestyle\endcsname
\providecommand{\newblock}{\relax}
\providecommand{\bibinfo}[2]{#2}
\providecommand{\BIBentrySTDinterwordspacing}{\spaceskip=0pt\relax}
\providecommand{\BIBentryALTinterwordstretchfactor}{4}
\providecommand{\BIBentryALTinterwordspacing}{\spaceskip=\fontdimen2\font plus
\BIBentryALTinterwordstretchfactor\fontdimen3\font minus
  \fontdimen4\font\relax}
\providecommand{\BIBforeignlanguage}[2]{{%
\expandafter\ifx\csname l@#1\endcsname\relax
\typeout{** WARNING: IEEEtran.bst: No hyphenation pattern has been}%
\typeout{** loaded for the language `#1'. Using the pattern for}%
\typeout{** the default language instead.}%
\else
\language=\csname l@#1\endcsname
\fi
#2}}
\providecommand{\BIBdecl}{\relax}
\BIBdecl

\bibitem{tai2017virtual}
L.~Tai, G.~Paolo, and M.~Liu, ``Virtual-to-real deep reinforcement learning:
  Continuous control of mobile robots for mapless navigation,'' in \emph{2017
  IEEE/RSJ International Conference on Intelligent Robots and Systems (IROS)},
  Sep 2017.

\bibitem{chen2017socially}
Y.~F. Chen, M.~Everett, M.~Liu, and J.~P. How, ``Socially aware motion planning
  with deep reinforcement learning,'' \emph{arXiv preprint arXiv:1703.08862},
  2017.

\bibitem{helbing1995social}
D.~Helbing and P.~Molnar, ``Social force model for pedestrian dynamics,''
  \emph{Physical review E}, vol.~51, no.~5, p. 4282, 1995.

\bibitem{fiorini1998motion}
P.~Fiorini and Z.~Shiller, ``Motion planning in dynamic environments using
  velocity obstacles,'' \emph{The International Journal of Robotics Research},
  vol.~17, no.~7, pp. 760--772, 1998.

\bibitem{van2008reciprocal}
J.~Van~den Berg, M.~Lin, and D.~Manocha, ``Reciprocal velocity obstacles for
  real-time multi-agent navigation,'' in \emph{Robotics and Automation, 2008.
  ICRA 2008. IEEE International Conference on}.\hskip 1em plus 0.5em minus
  0.4em\relax IEEE, 2008, pp. 1928--1935.

\bibitem{van2011reciprocal}
J.~Van Den~Berg, S.~Guy, M.~Lin, and D.~Manocha, ``Reciprocal n-body collision
  avoidance,'' \emph{Robotics research}, pp. 3--19, 2011.

\bibitem{alonso2013optimal}
J.~Alonso-Mora, A.~Breitenmoser, M.~Rufli, P.~Beardsley, and R.~Siegwart,
  ``Optimal reciprocal collision avoidance for multiple non-holonomic robots,''
  in \emph{Distributed Autonomous Robotic Systems}.\hskip 1em plus 0.5em minus
  0.4em\relax Springer, 2013, pp. 203--216.

\bibitem{snape2010smooth}
J.~Snape, J.~Van Den~Berg, S.~J. Guy, and D.~Manocha, ``Smooth and
  collision-free navigation for multiple robots under differential-drive
  constraints,'' in \emph{Intelligent Robots and Systems (IROS), 2010 IEEE/RSJ
  International Conference on}.\hskip 1em plus 0.5em minus 0.4em\relax IEEE,
  2010, pp. 4584--4589.

\bibitem{alonso2012reciprocal}
J.~Alonso-Mora, A.~Breitenmoser, P.~Beardsley, and R.~Siegwart, ``Reciprocal
  collision avoidance for multiple car-like robots,'' in \emph{Robotics and
  Automation (ICRA), 2012 IEEE International Conference on}.\hskip 1em plus
  0.5em minus 0.4em\relax IEEE, 2012, pp. 360--366.

\bibitem{wilkie2009generalized}
D.~Wilkie, J.~Van Den~Berg, and D.~Manocha, ``Generalized velocity obstacles,''
  in \emph{Intelligent Robots and Systems, 2009. IROS 2009. IEEE/RSJ
  International Conference on}.\hskip 1em plus 0.5em minus 0.4em\relax IEEE,
  2009, pp. 5573--5578.

\bibitem{kim2015brvo}
S.~Kim, S.~J. Guy, W.~Liu, D.~Wilkie, R.~W. Lau, M.~C. Lin, and D.~Manocha,
  ``Brvo: Predicting pedestrian trajectories using velocity-space reasoning,''
  \emph{The International Journal of Robotics Research}, vol.~34, no.~2, pp.
  201--217, 2015.

\bibitem{lavalle1998rapidly}
S.~M. LaValle, ``Rapidly-exploring random trees: A new tool for path
  planning,'' 1998.

\bibitem{kavraki1996probabilistic}
L.~E. Kavraki, P.~Svestka, J.-C. Latombe, and M.~H. Overmars, ``Probabilistic
  roadmaps for path planning in high-dimensional configuration spaces,''
  \emph{IEEE transactions on Robotics and Automation}, vol.~12, no.~4, pp.
  566--580, 1996.

\bibitem{karaman2011sampling}
S.~Karaman and E.~Frazzoli, ``Sampling-based algorithms for optimal motion
  planning,'' \emph{The international journal of robotics research}, vol.~30,
  no.~7, pp. 846--894, 2011.

\bibitem{hart1968formal}
P.~E. Hart, N.~J. Nilsson, and B.~Raphael, ``A formal basis for the heuristic
  determination of minimum cost paths,'' \emph{IEEE transactions on Systems
  Science and Cybernetics}, vol.~4, no.~2, pp. 100--107, 1968.

\bibitem{dijkstra1959note}
E.~W. Dijkstra, ``A note on two problems in connexion with graphs,''
  \emph{Numerische mathematik}, vol.~1, no.~1, pp. 269--271, 1959.

\bibitem{stentz1994optimal}
A.~Stentz, ``Optimal and efficient path planning for partially-known
  environments,'' in \emph{Robotics and Automation, 1994. Proceedings., 1994
  IEEE International Conference on}.\hskip 1em plus 0.5em minus 0.4em\relax
  IEEE, 1994, pp. 3310--3317.

\bibitem{fox1997dynamic}
D.~Fox, W.~Burgard, and S.~Thrun, ``The dynamic window approach to collision
  avoidance,'' \emph{IEEE Robotics \& Automation Magazine}, vol.~4, no.~1, pp.
  23--33, 1997.

\bibitem{borenstein1991vector}
J.~Borenstein and Y.~Koren, ``The vector field histogram-fast obstacle
  avoidance for mobile robots,'' \emph{IEEE transactions on robotics and
  automation}, vol.~7, no.~3, pp. 278--288, 1991.

\bibitem{gammell2014informed}
J.~D. Gammell, S.~S. Srinivasa, and T.~D. Barfoot, ``Informed rrt*: Optimal
  sampling-based path planning focused via direct sampling of an admissible
  ellipsoidal heuristic,'' in \emph{Intelligent Robots and Systems (IROS 2014),
  2014 IEEE/RSJ International Conference on}.\hskip 1em plus 0.5em minus
  0.4em\relax IEEE, 2014, pp. 2997--3004.

\bibitem{klancar2005mobile}
G.~Klancar, D.~Matko, and S.~Blazic, ``Mobile robot control on a reference
  path,'' in \emph{Intelligent Control, 2005. Proceedings of the 2005 IEEE
  International Symposium on, Mediterrean Conference on Control and
  Automation}.\hskip 1em plus 0.5em minus 0.4em\relax IEEE, 2005, pp.
  1343--1348.

\bibitem{rohmer2013v}
E.~Rohmer, S.~P. Singh, and M.~Freese, ``V-rep: A versatile and scalable robot
  simulation framework,'' in \emph{Intelligent Robots and Systems (IROS), 2013
  IEEE/RSJ International Conference on}.\hskip 1em plus 0.5em minus 0.4em\relax
  IEEE, 2013, pp. 1321--1326.

\end{thebibliography}
\end{document}